\documentclass[10pt]{article} % For LaTeX2e
\usepackage[preprint]{tmlr}

\usepackage{amsmath,amsfonts,bm}

\def\eqref#1{equation~\ref{#1}}
\def\1{\bm{1}}

\DeclareMathAlphabet{\mathsfit}{\encodingdefault}{\sfdefault}{m}{sl}
\SetMathAlphabet{\mathsfit}{bold}{\encodingdefault}{\sfdefault}{bx}{n}

\usepackage{amssymb}
\usepackage{bm}
\usepackage{graphicx}
\usepackage{hyperref}
\usepackage{url}
\usepackage{booktabs}
\usepackage{threeparttable}

\usepackage{booktabs}
\usepackage{threeparttable}

\title{Feature Interaction Modeling for Neural Operators}

\author{\name Quan GU \email guquan1839@163.com \\
        \addr Independent Researcher
      \AND
      \name Xiaoduo lI \email 2024001756@link.tyut.edu.cn\\
      \addr Taiyuan University of Technology
      \AND 
      \name Hongxia LIU \email liuhongxia@tyut.edu.cn\\
      \addr Taiyuan University of Technology}

\def\month{MM}  % Insert correct month for camera-ready version
\def\year{YYYY} % Insert correct year for camera-ready version
\def\openreview{\url{https://openreview.net/forum?id=XXXX}} % Insert correct link to OpenReview for camera-ready version

\begin{document}

\maketitle

\begin{abstract}
Despite the many variants of DeepONet that have been proposed, query-based operator networks still struggle with shock-dominated and low-viscosity PDEs, whose sharp moving discontinuities and slowly decaying solution spectra challenge finite-dimensional separable representations. In this work, we propose \emph{Feature Interaction Modeling Operator} (FM-Operator), a point-wise query neural operator that explicitly models feature construction and interactions between sensor observations and query coordinates. Our design is motivated by a reinterpretation of the canonical DeepONet aggregation through the lens of multiplicative interactions. Specifically, the branch--trunk inner product admits the equivalent form
\(\boldsymbol{b}(u)^\top \boldsymbol{\tau}(y)=\boldsymbol{1}^\top \operatorname{diag}(\boldsymbol{b}(u))\,\boldsymbol{\tau}(y)\),
revealing that the two representations interact only along corresponding latent dimensions and therefore constitute a diagonally constrained multiplicative interaction. This observation suggests that, beyond improving the individual branch and trunk networks, the structure through which function and query representations interact is itself an important inductive bias in point-wise operator learning. FM-Operator accordingly redesigns both feature construction and feature interaction, enabling structured information exchange beyond the conventional branch--trunk coupling while retaining point-wise query evaluation. Experiments across multiple PDE benchmarks demonstrate that FM-Operator consistently outperforms vanilla DeepONet and achieves clear improvements over the strong Shift-DeepONet baseline, particularly in regimes involving shocks and strong discontinuities. These results suggest that explicitly designing representation construction and interaction provides a promising direction for improving the effectiveness of DeepONet-style query-based neural operators.
\end{abstract}

\section{Introduction}

Partial differential equations (PDEs) constitute a fundamental mathematical language for describing physical phenomena and play an indispensable role across science and engineering. Classical numerical solvers, including finite difference, finite volume, and finite element methods, have achieved remarkable success in the simulation of PDE-governed systems. However, repeatedly solving parameterized PDEs at high spatial and temporal resolutions can remain computationally expensive, particularly in applications requiring many-query evaluations or real-time prediction. In recent years, learning-based approaches have emerged as promising alternatives. Physics-informed neural networks (PINNs) incorporate governing equations into the optimization objective to approximate individual PDE solutions, whereas neural operators aim to learn mappings directly between function spaces, thereby amortizing the computational cost over a family of PDE instances.

From an architectural perspective, existing neural operator methods can be roughly viewed through several representative paradigms. One prominent family follows a \emph{point-wise query} formulation, exemplified by the Deep Operator Network (DeepONet) and its variants, where an input function is encoded from sensor observations and the target solution is evaluated at arbitrary query coordinates. Another family adopts a \emph{field-to-field} formulation through learned nonlocal operators, including the Fourier Neural Operator (FNO), Graph Neural Operator (GNO), and their variants. More recently, Transformer-based operator models have introduced attention-based global interactions among spatial, temporal, and physical representations, and have been increasingly explored in heterogeneous, multi-physics, and multi-operator settings. Rather than constituting a strict taxonomy, these paradigms highlight different choices for representing functions and exchanging information within an operator network.

Among these approaches, DeepONet provides one of the most influential formulations of point-wise operator learning. Given sensor observations of an input function $u$ and a query coordinate $y$, the canonical DeepONet approximates an operator $\mathcal{G}$ as
\begin{equation}
    \mathcal{G}_{\theta}(u)(y)
    =
    \sum_{k=1}^{p} b_k(u)\tau_k(y)
    =
    \boldsymbol{b}(u)^{\top}\boldsymbol{\tau}(y),
    \label{eq:deeponet}
\end{equation}
where $\boldsymbol{b}(u)\in\mathbb{R}^{p}$ denotes the output of the branch network and
$\boldsymbol{\tau}(y)\in\mathbb{R}^{p}$ denotes the output of the trunk network. This formulation can be interpreted as a finite-dimensional separable reconstruction of the target function, in which the trunk network provides a learned set of basis functions and the branch network predicts their input-dependent coefficients.

Theoretical analysis by ~\cite{10.1093/imatrm/tnac001} reveals an important limitation of such linear reconstruction. They decompose the DeepONet error into encoding, approximation, and reconstruction components and show that the reconstruction error is fundamentally controlled by the spectral decay of the covariance operator associated with the push-forward measure $\mathcal{G}_{\#}\mu$. For a reconstruction space of dimension $p$, their analysis gives
\begin{equation}
    \left(
        \sum_{k>p}\lambda_k
    \right)^{1/2}
    \leq
    \widehat{\mathcal{E}}_{\mathcal{R}}
    \leq
    \widehat{\mathcal{E}},
    \label{eq:deeponet_lower_bound}
\end{equation}
where $\{\lambda_k\}_{k\geq 1}$ are the ordered eigenvalues of the covariance operator of $\mathcal{G}_{\#}\mu$, $\widehat{\mathcal{E}}_{\mathcal{R}}$ denotes the reconstruction error, and $\widehat{\mathcal{E}}$ denotes the total DeepONet approximation error. Consequently, slowly decaying output spectra lead to slowly decaying error lower bounds, regardless of the expressive power of the branch network. For a representative inviscid Burgers problem with moving shocks, Lanthaler et al. further show that
$\lambda_k\sim k^{-2}$, yielding a lower bound of order
$\widehat{\mathcal{E}}\gtrsim p^{-1/2}$. This result illustrates why a fixed finite-dimensional linear reconstruction can become inefficient when the solution family contains moving discontinuities.

This limitation has motivated several extensions of DeepONet that replace or relax its fixed linear reconstruction. Shift-DeepONet proposed by ~\cite{lanthaler2022nonlinearreconstructionoperatorlearning} introduces input-dependent scale and shift transformations into the query coordinates and takes the form
\begin{equation}
    \mathcal{G}_{\theta}(u)(y)
    =
    \sum_{k=1}^{p}
    b_k(u)\,
    \tau_k\!\left(
        A_k(u)y+\gamma_k(u)
    \right),
    \label{eq:shift_deeponet}
\end{equation}
where $A_k(u)$ and $\gamma_k(u)$ are generated by additional scale and shift networks, respectively. By allowing the effective basis functions to depend on the input function, Shift-DeepONet can represent translated or rescaled sharp structures with substantially fewer modes than a fixed linear reconstruction. Related approaches, such as FlexDeepONet\citep{VENTURI2023115718} and NOMAD\citep{3600270.3600675}, further relax the reconstruction structure through input-dependent coordinate transformations or nonlinear decoders.

These developments primarily address the \emph{reconstruction} side of DeepONet. In this work, we investigate a complementary question: \emph{how should the different sources of information in a point-wise operator be represented and interact before producing the final prediction?} Our motivation originates from a simple methodological analogy between low-rank reduced-order modeling and factorization-based predictive models. Proper Orthogonal Decomposition (POD) represents high-dimensional solution fields through a small number of latent modes, while Matrix Factorization (MF) represents relations between entities through low-dimensional latent factors. Although the two methods arise from different mathematical settings and optimization objectives, both exploit low-rank latent representations to capture dominant structure.

An instructive development in recommender systems is the Factorization Machine (FM)~\citep{10.1109/ICDM.2010.127}. Matrix factorization models interactions through inner products between latent representations, whereas FM generalizes the factorization principle to arbitrary feature vectors and explicitly models pairwise feature interactions using factorized parameters. In particular, conventional matrix factorization can be recovered as a special case of FM under an appropriate feature encoding. This transition from \emph{latent representation} to \emph{explicit interaction modeling} motivates us to reconsider the standard DeepONet formulation from the same perspective: after function and query information have been embedded into latent features, is a single inner product necessarily the most appropriate mechanism through which these representations should communicate?

A closely related interpretation is provided by HyperDeepONet~\citep{lee2023hyperdeeponet}. Rather than viewing DeepONet solely as a basis expansion, HyperDeepONet interprets operator architectures according to how information from the input function $u$ is injected into a query-dependent target network. Under this view, the trunk network of vanilla DeepONet can be regarded as the target network up to its final output layer, while the branch output $\boldsymbol{b}(u)$ provides the input-dependent weights connecting the last hidden representation to the scalar prediction. Shift-DeepONet and FlexDeepONet can similarly be interpreted as conditioning selected parameters of the target network, and HyperDeepONet generalizes this idea by allowing a hypernetwork to generate a larger subset of target-network parameters. NOMAD is also analyzed within this broader discussion of input-conditioned target networks, although it employs a different nonlinear decoder structure.

This hypernetwork interpretation reveals that the vanilla DeepONet aggregation can also be examined through the framework of multiplicative interactions. \cite{Jayakumar2020Multiplicative} organize multiplicative neural interactions according to the structure of the input-dependent projection, ranging from scalar scaling to diagonal modulation and general bilinear transformations. The standard DeepONet interaction in Eq.~\eqref{eq:deeponet} admits the exact rewriting
\begin{equation}
    \boldsymbol{b}(u)^{\top}\boldsymbol{\tau}(y)
    =
    \boldsymbol{1}^{\top}
    \operatorname{diag}\!\big(\boldsymbol{b}(u)\big)
    \boldsymbol{\tau}(y),
    \label{eq:diagonal_deeponet}
\end{equation}
which exposes the branch output as the diagonal entries of an input-dependent linear transformation applied to the trunk representation. Equivalently,
\begin{equation}
    \boldsymbol{b}(u)^{\top}\boldsymbol{\tau}(y)
    =
    \sum_{i=1}^{p}
    b_i(u)\tau_i(y),
\end{equation}
so the explicit branch--trunk interaction occurs only between corresponding latent dimensions. Under the multiplicative-interaction perspective, vanilla DeepONet can therefore be interpreted as employing a \emph{diagonally constrained multiplicative interaction}.

This observation should not be confused with a limitation of the universal approximation capability of DeepONet. Both the branch and trunk networks may individually construct highly nonlinear and expressive representations. Rather, Eq.~\eqref{eq:diagonal_deeponet} identifies a structural constraint on the \emph{interface} through which the two representations exchange information: after potentially rich nonlinear feature extraction, their explicit interaction is reduced to dimension-wise multiplication followed by summation. We therefore hypothesize that the construction of operator features and the structure of their interactions constitute an additional source of inductive bias that has received comparatively less attention than the design of the individual branch and trunk networks.

\paragraph{Our approach:}
Motivated by this perspective, we propose the \emph{Feature Interaction Modeling Operator} (FM-Operator), a point-wise query neural operator that treats \emph{feature construction} and \emph{feature interaction} as explicit architectural design dimensions. Instead of relying exclusively on the conventional branch--trunk inner product, FM-Operator constructs structured representations from sensor observations and query coordinates and introduces interaction mechanisms that facilitate richer information exchange among these representations. Importantly, our objective is not to replace the diagonal interaction with an unrestricted high-order interaction merely for additional model capacity. Rather, we seek structured interaction mechanisms that provide an effective inductive bias while retaining the key advantages of point-wise query operators, including continuous-coordinate evaluation and flexible resolution querying.

We evaluate FM-Operator on multiple PDE benchmarks, with particular emphasis on low-viscosity and shock-dominated regimes containing sharp gradients and moving discontinuities. The results show that FM-Operator consistently improves upon vanilla DeepONet and achieves clear gains over the strong Shift-DeepONet baseline on these challenging problems. These findings indicate that the performance of DeepONet-style operators is influenced not only by the expressiveness of their branch and trunk subnetworks or by the adaptivity of their reconstruction basis, but also by how heterogeneous operator features are constructed and how their interactions are parameterized.

\begin{figure}[htbp]
    \centering
    \includegraphics[width=\linewidth]{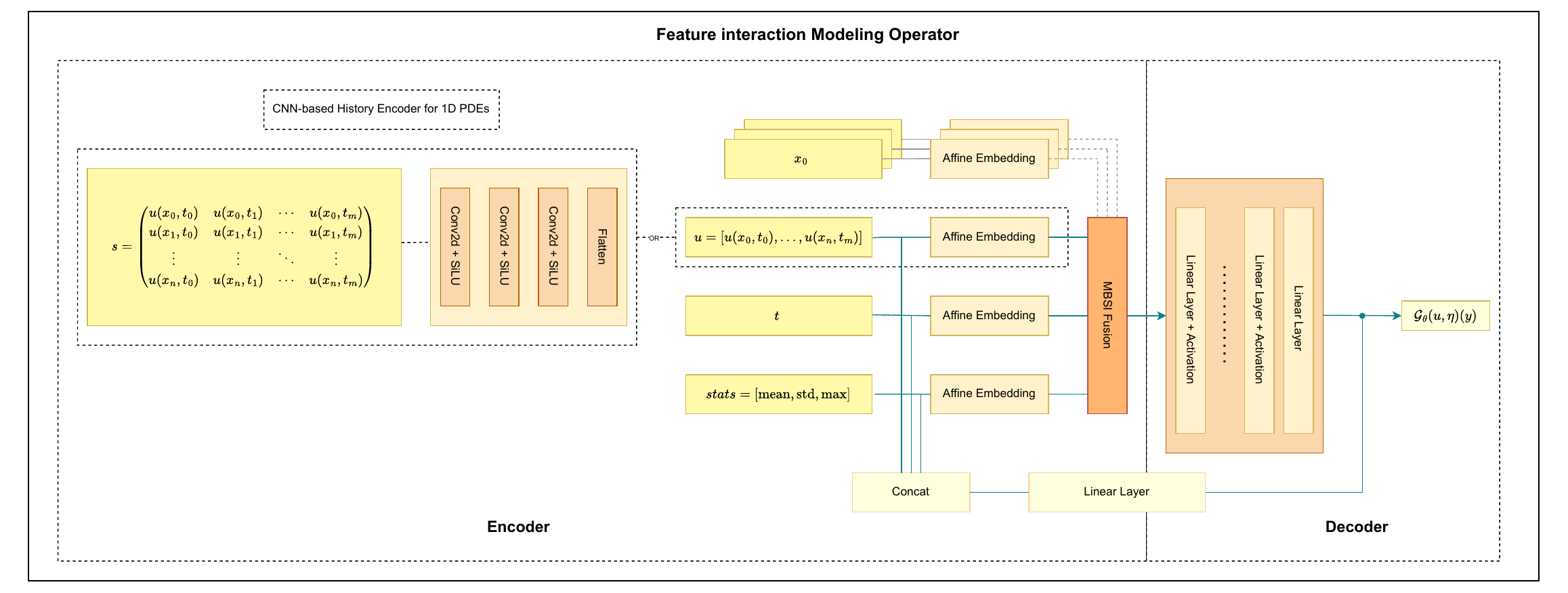}
    \caption{Overview of the Feature Interaction Modeling Operator(FM-Operator with MBSI Interaction).}
    \label{fig:fm-operator}
\end{figure}

\paragraph{Our contributions are summarized as follows:}
\begin{itemize}
    \item We reinterpret the canonical DeepONet aggregation from the perspective of multiplicative interactions and show that its branch--trunk inner product is exactly expressible as an input-dependent diagonal transformation. This exposes feature interaction as an explicit architectural design dimension in point-wise operator learning.

    \item Motivated by the connection between low-rank factorization and structured interaction modeling, As Figure \ref{fig:fm-operator} depicted, we propose FM-Operator, which jointly redesigns feature construction and feature interaction while retaining the point-wise query formulation of DeepONet.

    \item We conduct experiments across multiple PDE benchmarks, including low-viscosity and shock-dominated problems, and demonstrate consistent improvements over vanilla DeepONet and strong performance relative to Shift-DeepONet, highlighting the effectiveness of interaction-aware inductive biases for challenging operator-learning problems.
\end{itemize}

\section{Related Work}

\subsection{Deep Operator Network}

DeepONet learns nonlinear operators through a branch network that encodes an input function sampled at a set of sensors and a trunk network that encodes the query coordinates. Their latent representations are combined by an inner product to reconstruct the output function \citep{Lu_2021}. POD-DeepONet replaces the learned trunk basis with precomputed Proper Orthogonal Decomposition (POD) modes, leaving the branch network to predict the corresponding coefficients \citep{Lu_2022}. For operators with multiple input functions, MIONet extends the branch--trunk architecture to product spaces using multiple branch networks and tensor-product interactions \citep{doi:10.1137/22M1477751}.

The linear reconstruction used by standard DeepONet can become inefficient for transport-dominated problems with moving discontinuities. Lanthaler et al.\ established approximation lower bounds for operator learners with linear reconstruction \citep{10.1093/imatrm/tnac001}. Building on this analysis, Hadorn proposed Shift-DeepONet, where the trunk representation is transformed by input-dependent shift and scale functions \citep{lanthaler2022nonlinearreconstructionoperatorlearning}. Going further, NOMAD replaces the linear basis expansion with a nonlinear decoder that jointly takes the latent representation of the input function and the query location. HyperDeepONet instead uses a hypernetwork to generate the parameters of a target network from the input function, allowing the reconstruction network itself to depend on the operator input \citep{lee2023hyperdeeponet}.

DeepONet has also been extended to varying geometries. Geom-DeepONet incorporates point-cloud representations and signed distance information for field prediction on parameterized three-dimensional geometries \citep{HE2024117130}. Point-DeepONet further integrates PointNet into the DeepONet framework to encode non-parametric point-cloud geometries and variable loading conditions without explicit geometric parameterization \cite{PARK2026108560}.

\subsection{Fourier and Geometry-Aware Neural Operators}

Fourier Neural Operator (FNO) parameterizes the kernel of a neural integral operator in Fourier space, enabling efficient global information mixing on regular grids \citep{li2021fourier}. In contrast to the query-wise evaluation of DeepONet, FNO updates the solution field through repeated spectral operator layers and has shown strong performance on PDE benchmarks including Burgers' equation, Darcy flow, and Navier--Stokes equations.

Physics-Informed Neural Operator (PINO) augments FNO with PDE constraints during training, allowing solution data and governing equations to be imposed at different resolutions \citep{10.1145/3648506}. This formulation improves physical consistency and supports zero-shot super-resolution beyond the discretization used for the training data.

For PDEs on irregular and varying domains, Geometry-Informed Neural Operator (GINO) combines graph-based integral operators with Fourier layers \citep{3666122.3667678}. Irregular geometric inputs are mapped to a regular latent grid, processed by FNO layers, and projected back to the original geometry. Point Cloud Neural Operator (PCNO) instead defines integral and differential operators directly on point clouds, allowing a common operator formulation to handle complex geometries and different discretizations \cite{ZENG2025118022}.

\subsection{Factorized Feature Interaction}

Factorization Machines (FM) model second-order feature interactions through low-dimensional latent factors rather than assigning an independent parameter to every feature pair \citep{10.1109/ICDM.2010.127}. For feature embeddings $\{\boldsymbol{e}_i\}_{i=1}^{n}$, the interaction between two fields is represented through the inner product of their latent factors, providing a parameter-efficient representation of pairwise dependencies.

Neural Factorization Machines (NFM) combine these factorized interactions with nonlinear neural representations \citep{10.1145/3077136.3080777}. Its bi-interaction pooling layer aggregates pairwise element-wise products as

\begin{equation}
    \boldsymbol{z}_{\boldsymbol{BI}}
    =
    \sum_{i<j}\boldsymbol{e}_i\odot\boldsymbol{e}_j
    =
    \frac{1}{2}
    \left[
    \left(\sum_i\boldsymbol{e}_i\right)
    \odot
    \left(\sum_i\boldsymbol{e}_i\right)
    -
    \sum_i\boldsymbol{e}_i\odot\boldsymbol{e}_i
    \right],
\end{equation}

where $\odot$ denotes the Hadamard product. The resulting interaction vector is then processed by nonlinear layers, retaining the explicit second-order structure of FM while enabling higher-order nonlinear modeling.

Later factorization models modify either the importance or the order of these interactions. AFM assigns attention weights to individual pairwise interactions instead of treating them uniformly \citep{3172077.3172324}. DeepFM combines an FM component for explicit low-order interactions with a deep network for implicit high-order interactions \citep{3172077.3172127}, while xDeepFM introduces the Compressed Interaction Network (CIN) to explicitly construct bounded-degree higher-order interactions at the vector level \citep{10.1145/3219819.3220023}.

A related line of work models interactions through bilinear and tensor fusion. A full bilinear interaction uses the outer product of two representations and therefore retains cross-dimensional feature interactions, but its dimensionality grows quadratically with the feature size. Low-rank bilinear pooling factorizes the bilinear transformation into learned projections followed by a Hadamard product \citep{kim2017hadamard}. In this form, projected element-wise multiplication provides an efficient low-rank representation of the full bilinear interaction. MUTAN extends bilinear fusion using Tucker decomposition of the interaction tensor, controlling model complexity while retaining richer cross-dimensional dependencies \citep{Ben-younes_2017_ICCV}.

These interaction mechanisms have mainly been developed in recommendation and multimodal learning. DeepONet-style neural operators, including variants with nonlinear or input-dependent reconstruction, primarily modify the encoder, basis, or decoder while using comparatively fixed mechanisms to combine input-function and query representations. In this work, we instead treat this fusion explicitly as a feature-interaction problem and introduce factorized interactions between operator inputs and query variables.

\section{A Unified View of DeepONets via HyperNetworks and Multiplicative Interactions}
\label{sec:unified}

\subsection{Preliminaries: DeepONet}
DeepONet \citep{Lu_2021} is the first neural operator architecture designed to learn nonlinear mappings between function spaces.
Given an input function $u\in\mathcal{U}$ and a nonlinear operator
\begin{equation}
    \mathcal{G}:\mathcal{U}\rightarrow\mathcal{V},
\end{equation}
DeepONet represents the input function through a branch network and the query location $y$ through a trunk network.
For measurements of $u$ at $m$ fixed sensors,
\begin{equation}
    u
    =
    \left[
    u(s_1),u(s_2),\ldots,u(s_m)
    \right]^{\top},
\end{equation}
the branch and trunk networks produce
\begin{equation}
    \boldsymbol{b}(u)
    =
    \left[
    b_1(u),\ldots,b_p(u)
    \right]^{\top},
    \qquad
    \boldsymbol{\tau}(y)
    =
    \left[
    \tau_1(y),\ldots,\tau_p(y)
    \right]^{\top}.
\end{equation}
The operator value at the query $y$ is then approximated by
\begin{equation}
    \widehat{\mathcal{G}}_{\theta}(u)(y)
    =
    \sum_{k=1}^{p}
    b_k(u)\tau_k(y)
    +
    b_0
    =
    \boldsymbol{b}(u)^{\top}
    \boldsymbol{\tau}(y)
    +
    b_0.
    \label{eq:deeponet}
\end{equation}

\subsection{HyperNetworks and Multiplicative Interactions}
~\cite{lee2023hyperdeeponet} revisited DeepONet and its variants from the perspective of target-network parameterization.
They think the DeepONet and its variants can be distinguished by how information from the input function \(u\)
is injected in to the target network \(G_\theta(u)\).
The parameter-injection perspective has a direct algebraic connection to the analysis of multiplicative interactions by ~\cite{Jayakumar2020Multiplicative}.
Given two representations, 
\begin{equation}
    x\in\mathbb R^{n},
    \qquad
    z\in\mathbb R^{m},
\end{equation}
We aim to approximate an unknown function that captures the multiplicative interaction between them through a neural network \(f\). Therefore, a general multiplicative interaction can be written as
\begin{equation}
    f(x,z)
    =
    z^{\top}\mathbb W x
    +
    z^{\top}U
    +
    Vx
    +
    b,
    \label{eq:mult-gen}
\end{equation}
where $\mathbb W\in\mathbb R^{m\times n\times k}$ is a third-order interaction tensor.

Hypernetwork can be thought as a neural nework \(g\) used to generate the weights \(z\) of main network, hence, the \(f(x;\theta)\) becomes 
\(f(x;g(z;\phi))\). This type of network is equivalent to the multiplicative interaction.

We can define the conditional weight matrix and bias as
\begin{equation}
    W^{'}(z)=z^{\top}\mathbb W+V,
    \qquad
    b^{'}(z)=z^{\top}U+b,
    \label{eq:cond-param}
\end{equation}
equation \eqref{eq:mult-gen} can be rewritten as
\begin{equation}
    f(x;g(z;\phi))=W^{'}(z)\,x+b^{'}(z).
    \label{eq:mult-as-hyper}
\end{equation}

Thus, when the conditional generation of target-network parameters is an affine map of the conditioning representation, hypernetworks and multiplicative interactions are algebraically equivalent.
The conditioning representation $z$ and the target representation $x$ are not combined by simple concatenation; rather, they interact multiplicatively through the linear transformation $W^{'}(z)$ determined by $z$.

\subsection{Vanilla DeepONet as a Restricted Multiplicative Interaction}

To apply this formalism to vanilla DeepONet, the \eqref{eq:mult-gen} becomes
\begin{equation}
    \tau(y; b(z;\phi)) = W^{'}(\boldsymbol{b}(u))\,\boldsymbol{\tau}(y)+b^{'}(\boldsymbol{b}(u)) =
    \boldsymbol{b}(u)^{\top}\mathbb W \boldsymbol{\tau}(y)
    +
    \boldsymbol{b}(u)^{\top}U
    +
    V\boldsymbol{\tau}(y)
    +
    b,
    \label{eq:vanilla-don}
\end{equation}
and specialize \eqref{eq:mult-gen} to scalar output $k=1$ with
\begin{equation}
    \mathbb W=I_p,
    \qquad
    U=0,
    \qquad
    V=0.
\end{equation}
Then \eqref{eq:vanilla-don} becomes
\begin{equation}
    \tau(y; b(z;\phi)) 
    =
    \boldsymbol{b}(u)^{\top}\boldsymbol{\tau}(y)+b,
    \label{eq:deeponet-mult}
\end{equation}
which exactly recovers vanilla DeepONet.
Equivalently,
\begin{equation}
    \boldsymbol{b}(u)^{\top}\boldsymbol{\tau}(y)
    =
    \boldsymbol 1^{\top}
    \left[
    \boldsymbol{b}(u)\odot
    \boldsymbol{\tau}(y)
    \right].
    \label{eq:hadamard}
\end{equation}

Therefore, the branch--trunk structure of vanilla DeepONet can be interpreted not only as a coefficient--basis expansion but also as a restricted multiplicative interaction: the input-function representation and the query representation first undergo dimension-wise multiplicative interaction and are then contracted to a scalar output by a fixed summation map.

\subsection{A General Representation--Interaction--Decoder Framework}
Based on this observation, we abstract point-wise query operator learning into three independent components:
\begin{equation}
    z_u=E_u(u),
    \qquad
    z_y=E_y(y),
    \label{eq:enc}
\end{equation}
\begin{equation}
    h=\Phi(z_u,z_y),
    \label{eq:interact}
\end{equation}
\begin{equation}
    \widehat{\mathcal G}_{\theta}(u)(y)
    =
    D(h),
    \label{eq:dec}
\end{equation}
or, in compact form,
\begin{equation}
    \boxed{
    \widehat{\mathcal G}_{\theta}(u)(y)
    =
    D\!\left(
    \Phi\!\left(E_u(u),E_y(y)\right)
    \right).
    }
    \label{eq:rid}
\end{equation}

Here $E_u$ and $E_y$ define the representation construction for the input function and the query variable, respectively; $\Phi$ defines the conditional interaction between the two representations; and $D$ decodes the interaction representation into the target quantity.
From this unified perspective, the core architectural design of point-wise query neural operators reduces to two fundamental questions:

\begin{itemize}
    \item[(i)] \textbf{Representation Construction:}
    \( z_u=E_u(u), \ z_y=E_y(y), \)
    i.e., how to construct feature representations that characterize the input function, the query variable, and any necessary geometric information. Methods such as Geo-DeepONet and Point-DeepONet mainly improve this component by incorporating geometric descriptions or point cloud information into the representation space.

    \item[(ii)] \textbf{Representation Interaction:}
    \( h=\Phi(z_u,z_y), \)
    i.e., in what form the information of the input function acts on the query representation to jointly determine the output. Vanilla DeepONet accomplishes this interaction through a fixed inner product, whereas Shift-DeepONet, NOMAD, and HyperDeepONet introduce different forms of conditional dependence via input-dependent coordinate transformations, joint nonlinear decoding, and target-network parameter generation, respectively.
\end{itemize}

Consequently, under the unified representation--interaction--decoder framework, the architectural differences among point-wise query neural operators reduce to two interrelated design degrees of freedom: \emph{how to construct the representations, and how these representations interact}.

\section{Feature Interaction Modeling Operator}
\subsection{Architecture of Feature Interaction Modeling Operator}

\paragraph{Representation in Feature interaction Modeling Operator}We propose the \emph{Feature Interaction Modeling Operator} (FM-Operator), a query-wise neural operator that explicitly models interactions among different components of the operator input before nonlinear reconstruction. Given a sampled input history and a query location, FM-Operator first encodes heterogeneous input components into a common latent space, performs structured pairwise interactions among them, and then reconstructs the target value through a nonlinear decoder.

For each query, we construct four feature groups,
\begin{equation}
    \mathcal{X}
    =
    \left\{
        \boldsymbol{s}, \boldsymbol{x}, t, \boldsymbol{c}
    \right\},
\end{equation}
where $\boldsymbol{s}$ denotes the sampled sensor/history observations, $x\in\mathbb{R}$ and $t\in\mathbb{R}$ denote the spatial and temporal query coordinates, respectively, and $\boldsymbol{c}\in\mathbb{R}^{3}$ denotes a low-dimensional statistical representation derived from the observed input. Although $\boldsymbol{c}$ is computed from the sensor observations, it is embedded separately to expose global statistical information explicitly to the subsequent interaction module rather than requiring it to be recovered implicitly from the high-dimensional history representation.

Each feature group is independently mapped into a shared $d$-dimensional latent space:
\begin{equation}
    \boldsymbol{e}_i = E_i(\boldsymbol{x}_i),
    \qquad
    \boldsymbol{e}_i \in \mathbb{R}^{d},
    \qquad i=0,\ldots,3,
\label{eq:field_embedding}
\end{equation}
The history observations are encoded by a dedicated history encoder, while the spatial coordinate, temporal coordinate, and statistical features are embedded by separate learnable projections. This separation preserves the semantic roles of the different input components before interaction.

\paragraph{Multi-head Bilinear Symmetric Interaction}Given the embeddings
\begin{equation}
    \left\{
        \boldsymbol{e}_0,
        \boldsymbol{e}_1,
        \boldsymbol{e}_2,
        \boldsymbol{e}_3
    \right\},
\end{equation}
FM-Operator constructs pairwise interactions using $H$ independent interaction heads. For the $h$-th head, every feature group is associated with two learnable projections,
\begin{equation}
    L_i^{(h)},R_i^{(h)}
    \in
    \mathbb{R}^{r\times d},
\end{equation}
where $r$ denotes the interaction width. The interaction between two feature groups $\boldsymbol{e}_i$ and $\boldsymbol{e}_j$ is defined as
\begin{equation}
\begin{aligned}
    \boldsymbol{\phi}_{ij}^{(h)}
    =&
    \left(
        L_i^{(h)}\boldsymbol{e}_i
    \right)
    \odot
    \left(
        R_j^{(h)}\boldsymbol{e}_j
    \right)
    \\
    &+
    \left(
        L_j^{(h)}\boldsymbol{e}_j
    \right)
    \odot
    \left(
        R_i^{(h)}\boldsymbol{e}_i
    \right),
\end{aligned}
\label{eq:fm_pairwise}
\end{equation}
where $\odot$ denotes the Hadamard product.

All pairwise interactions within the same head are aggregated by element-wise summation,
\begin{equation}
    \boldsymbol{p}^{(h)}
    =
    \sum_{i<j}
    \boldsymbol{\phi}_{ij}^{(h)}
    \in
    \mathbb{R}^{r}.
\label{eq:fm_head_pool}
\end{equation}
For four feature groups, this results in six explicit pairwise interactions. The use of pairwise pooling follows the factorization-based interaction principle while maintaining a compact interaction representation.

The outputs of all heads are concatenated and projected back to the latent dimension:
\begin{equation}
    \boldsymbol{p}
    =
    \sigma
    \left(
        W_o
        \left[
            \boldsymbol{p}^{(1)};
            \boldsymbol{p}^{(2)};
            \ldots;
            \boldsymbol{p}^{(H)}
        \right]
        +
        \boldsymbol{b}_o
    \right),
\label{eq:fm_multihead}
\end{equation}

The fused interaction representation is subsequently decoded by a nonlinear multilayer perceptron,
\begin{equation}
    \hat{u}_{\boldsymbol{deep}}
    =
    D_{\theta}(\boldsymbol{p}),
\end{equation}
where $D_{\theta}$ consists of two hidden layers with SiLU activations.

In addition to the interaction branch, FM-Operator includes a direct linear path from the raw inputs to the output:
\begin{equation}
    \hat{u}_{\boldsymbol{wide}}
    =
    W_{\boldsymbol{wide}}
    \left[
        \boldsymbol{s};
        x;
        t;
        \boldsymbol{c}
    \right]
    +
    b_{\boldsymbol{wide}}.
\label{eq:fm_wide}
\end{equation}
The final prediction is obtained by combining the nonlinear interaction branch and the linear residual branch:
\begin{equation}
    \hat{u}
    =
    \hat{u}_{\boldsymbol{deep}}
    +
    \hat{u}_{\boldsymbol{wide}}.
\label{eq:fm_output}
\end{equation}
The wide branch directly captures first-order dependencies in the original input space, while the deep branch models nonlinear relationships constructed from explicit cross-group interactions.

\subsection{Structured Bilinear Feature Interaction}

Although Eq.~\eqref{eq:fm_pairwise} is implemented using projected Hadamard products, the resulting operation represents a structured bilinear interaction between different feature groups.

Let $\boldsymbol{l}_{i,h,k}^{\top}$ and $\boldsymbol{r}_{i,h,k}^{\top}$ denote the $k$-th rows of $L_i^{(h)}$ and $R_i^{(h)}$, respectively. The $k$-th component of the interaction between $\boldsymbol{e}_i$ and $\boldsymbol{e}_j$ can then be written as
\begin{equation}
\begin{aligned}
    \phi_{ij,h,k}
    =&
    \left(
        \boldsymbol{l}_{i,h,k}^{\top}\boldsymbol{e}_i
    \right)
    \left(
        \boldsymbol{r}_{j,h,k}^{\top}\boldsymbol{e}_j
    \right)
    \\
    &+
    \left(
        \boldsymbol{r}_{i,h,k}^{\top}\boldsymbol{e}_i
    \right)
    \left(
        \boldsymbol{l}_{j,h,k}^{\top}\boldsymbol{e}_j
    \right).
\end{aligned}
\label{eq:fm_bilinear_expansion}
\end{equation}
Equivalently,
\begin{equation}
    \phi_{ij,h,k}
    =
    \boldsymbol{e}_i^{\top}
    W_{ij,h,k}
    \boldsymbol{e}_j,
\label{eq:fm_bilinear_form}
\end{equation}
where
\begin{equation}
    W_{ij,h,k}
    =
    \boldsymbol{l}_{i,h,k}
    \boldsymbol{r}_{j,h,k}^{\top}
    +
    \boldsymbol{r}_{i,h,k}
    \boldsymbol{l}_{j,h,k}^{\top}.
\label{eq:fm_bilinear_matrix}
\end{equation}

This formulation highlights an important distinction from a direct Hadamard interaction. A conventional element-wise product,
\begin{equation}
    \boldsymbol{e}_i\odot\boldsymbol{e}_j,
\end{equation}
only couples coordinates with matching indices in the original embedding space. In contrast, Eq.~\eqref{eq:fm_bilinear_form} first projects both embeddings into learned interaction subspaces. Expanding one projected product gives
\begin{equation}
\begin{aligned}
    &
    \left(
        \boldsymbol{l}_{i,h,k}^{\top}\boldsymbol{e}_i
    \right)
    \left(
        \boldsymbol{r}_{j,h,k}^{\top}\boldsymbol{e}_j
    \right)
    \\
    =&
    \sum_{m=1}^{d}
    \sum_{n=1}^{d}
    l_{i,h,k,m}
    r_{j,h,k,n}
    e_{i,m}e_{j,n},
\end{aligned}
\end{equation}
which contains interactions between different latent coordinates $m$ and $n$. Consequently, the proposed interaction module can capture cross-dimensional dependencies without explicitly constructing the full outer product
$\boldsymbol{e}_i\boldsymbol{e}_j^{\top}$.

The bidirectional construction in Eq.~\eqref{eq:fm_pairwise} further uses separate left and right projections while treating each unordered feature pair consistently:
\begin{equation}
    \boldsymbol{\phi}_{ij}^{(h)}
    =
    \boldsymbol{\phi}_{ji}^{(h)}.
\end{equation}
Each interaction channel is therefore parameterized by a structured bilinear matrix composed of two rank-one terms, while multiple channels and multiple heads provide a collection of complementary interaction subspaces.

Compared with directly concatenating all input embeddings and relying on an MLP to discover their relationships implicitly, FM-Operator exposes second-order cross-group dependencies before nonlinear reconstruction. The subsequent decoder is then responsible for composing these interaction features into the final operator prediction.

\section{3D Geometry Feature Interaction Modeling Opererator}
As mentioned above, the representation construct is one of the most important design. Therefore, in this section, we proposed the Geometry Feature Interaction Modeling Operator(GFM-Operator) through Feature engineering.
We construct 3 Feature Groups that are Sensor, Query position, Query time, where the Representation of Sensor is the most important.

\subsection{Sensor Embedding}
We denote the history data collected at the sensor locations by
$\boldsymbol{h}_i \in \mathbb{R}^{d \times t \times n}$, where $d$ is the
value dimension (the unit wall normal $\boldsymbol{n} \in \mathbb{R}^3$
concatenated with the wall shear stress vector $\boldsymbol{\tau}
\in \mathbb{R}^3$), $t$ is the number of time steps, and $n$ is the number
of sensors.

As the first step, the history data is encoded along the value dimension by
a shared MLP, which projects the $d$-dimensional input into a
higher-dimensional feature space while keeping each sensor and time step
independent. Next, a temporal CNN is applied along the time axis to capture
the temporal dynamics of each sensor, producing a compact token per sensor.
Finally, for a query position, we compute its Euclidean or surface-geodesic distances to all sensors, select the nearest $k$ sensors, and obtain the query history embedding by aggregating the corresponding sensor tokens with Gaussian spatial weights.
\begin{equation}
  w_j \;=\; \boldsymbol{softmax}_j\!\Big(
    -\tfrac{1}{2}\big(d_j / b\big)^2 
  \Big),
  \qquad
  b \;=\; \boldsymbol{mean}_{j=2..k}\big(d_{(j)}\big),
  \qquad
  z_h(x) \;=\; \sum_{j=1}^{k} w_j \, \boldsymbol{token}_j,
\end{equation}
where $d_{(j)}$ denotes the sorted nearest-neighbour distances and the
closest neighbour is excluded from the bandwidth estimate $b$.

\section{Experiments}
In this section, we compared the Feature Interaction Modeling Operator with a series of variants of DeepONet. In addition, we conducted experiments on the Aneumo 3D geometry dataset.
As Table~\ref{tab:networkset} depicted, the network configurations in the ex1 and ex2 are the same. In addition, Table \ref{tab:exsetting} shows the training configguration.

\begin{table}[t]
  \centering
  \caption{Network setting. $m$: branch-input dim;
  $d_y$: query coordinate dim.}
  \label{tab:networkset}
  \begin{tabular}{@{}l l l@{}}
    \toprule
    Model & Branch net & Target net \\
    \midrule
    DeepONet
      & $m{-}256{-}256{-}256{-}32$
      & $d_y{-}256{-}256{-}256{-}32$ \\
    Shift-DeepONet
      & $m{-}256{-}256{-}256{-}32$ 
      & $(d_y{\cdot}32){-}256{-}256{-}256{-}32$ \\
    POD-DeepONet
      & $m{-}128{-}128{-}32$ 
      & fixed POD basis (32 modes) \\
    \midrule
    FM-Operator& \multicolumn{2}{l}{
        Decoder $128{-}128{-}128{-}128{-}1$} \\
    \bottomrule
  \end{tabular}
\end{table}

\subsection{Experiments Settings}

\paragraph{EX1 benchmark.}
In experiment 1, we consider five nonlinear benchmark problems:
\begin{itemize}
    \item \textbf{Buckley--Leverett}: $u_t+\partial_x f(u)=0$, with
    $f(u)=\dfrac{u^2}{u^2+a(1-u)^2}$ and $a=0.5$, $t\in[0,0.6]$,
    periodic boundary conditions;
    \item \textbf{Cubic Conservation}: $u_t+\partial_x(u^3/3)=\nu u_{xx}$,
    with $\nu=0.002$, $t\in[0,1]$, periodic boundary conditions;
    \item \textbf{Kuramoto--Sivashinsky}: $u_t+uu_x+u_{xx}+u_{xxxx}=0$,
    on a periodic domain of length $L=22$, $t\in[0,1]$;
    \item \textbf{Square-pulse Advection}: $u_t+c\,u_x=0$, with $c=0.6$,
    $t\in[0,1]$, periodic boundary conditions;
    \item \textbf{Periodic Burgers}: $u_t+uu_x=\nu u_{xx}$, with
    $\nu=0.001$, $t\in[0,1]$, periodic boundary conditions;
\end{itemize}

\begin{table}[htbp]
\centering
\caption{Experimental Settings}

\label{tab:exsetting}
\begin{tabular}{|l|c|c|c|}
\hline
\textbf{Item} & \textbf{EX1} & \textbf{EX2} & \textbf{EX3} \\ \hline
Data Scale & 256 train / 64 test & 9000 train / 1000 test & 647 train / 81 test \\ \hline
Validation Set & 64 & None & 81 \\ \hline
Sensors & 64 & 256 & 8000 \\ \hline
Input Dimensions & $20\times64=1280$ & $20\times256=5120$ & $20\times8000\times3=480000$ \\ \hline
Evaluation Points & $20\times128=2560$ &$20\times256 = 5120$ & $20\times13902=278040$ per case \\ \hline
Training Steps & 30000 & 50000 & $647\times20=12940$ \\ \hline
\end{tabular}
\end{table}

% ============ 方法/数据集章节片段 ============
\paragraph{EX2 benchmark.}
In experiment 2, we consider two nonlinear benchmark problems from the
PDEBench dataset \citep{NEURIPS2022_0a974713}:
\begin{itemize}
    \item \textbf{1D viscous Burgers}: $u_t+uu_x=\nu u_{xx}$, with
    $\nu=0.001$, $x\in[0,1]$, $t\in[0,1]$, periodic boundary
    conditions, taken from \texttt{1D\_Burgers\_Sols\_Nu0.001.hdf5}.
    \item \textbf{1D compressible Euler (shock-tube / Riemann)}:
    $\partial_t\boldsymbol U+\partial_x\boldsymbol F=\boldsymbol 0$, with
    $\boldsymbol U=(\rho,\rho v,E)^{\top}$ and
    $\boldsymbol F=(\rho v,\rho v^2+p,v(E+p))^{\top}$, closed by the
    ideal-gas equation of state $p=(\gamma-1)(E-\tfrac12\rho v^2)$ with
    $\gamma=1.4$, numerical dissipation coefficients
    $\eta=\zeta=10^{-8}$, and shock-tube (Riemann) initial data
    (\texttt{1D\_CFD\_Shock\_Eta1.e-8\_Zeta1.e-8\_trans\_Train.hdf5});
    the density field $\rho(x,t)$ is used as the observed quantity.
\end{itemize}

\paragraph{EX3 benchmark.}
In Experiment 3, we address the task of forecasting the wall shear stress (WSS) vector field over the entire vessel wall, a geometry-conditioned spatio-temporal operator learning problem, using 809 registered intracranial aneurysm geometries from the AneuG-Flow dataset.
\citep{ding2026aneugflow}.

\subsection{Results Analysis}
\paragraph{EX1: Nonlinear Equations}

\begin{table*}[htbp]  % 双栏文档请使用 table*；单栏请改为 table
\centering
\tiny                       % 缩小字号
\setlength{\tabcolsep}{2pt}      % 缩小列间距
\begin{threeparttable}
\caption{Relative L2 error (mean $\pm$ std over 3 seeds) of FM-Operator variants and DeepONet variants}
\label{tab:fmop_all}
\begin{tabular}{lcccccccc}
\toprule
 & \multicolumn{5}{c}{FM-Operator} & Shift-DeepONet & Vanilla DeepONet & POD-DeepONet \\
\cmidrule(lr){2-6}
 & MBSI-CNNe & MBSI-Ae & NFM-CNNe & Concat-CNNe & Outer-CNNe & & & \\
\midrule
Buckley-Leverett & $0.1016\pm0.0052$ & $\boldsymbol{0.0845\pm0.0007}$ & $0.1140\pm0.0104$ & $0.1169\pm0.0103$ & $0.1105\pm0.0052$ & $0.1045\pm0.0024$ & $0.0990\pm0.0054$ & $0.1018\pm0.0015$ \\
Cubic Conservation & $0.1032\pm0.0032$ & $\boldsymbol{0.1019\pm0.0053}$ & $0.2102\pm0.0106$ & $0.1672\pm0.0524$ & $0.1741\pm0.0202$ & $0.2943\pm0.0493$ & $0.2114\pm0.0014$ & $0.2284\pm0.0113$ \\
Kuramoto-Sivashinsky & $\boldsymbol{0.0155\pm0.0015}$ & $0.0262\pm0.0064$ & $0.2755\pm0.0835$ & $0.0860\pm0.0576$ & $0.0290\pm0.0008$ & $0.2975\pm0.0302$ & $0.1324\pm0.0003$ & $0.1693\pm0.0081$ \\
Square Advection & $\boldsymbol{0.1690\pm0.0024}$ & $0.1872\pm0.0106$ & $0.1804\pm0.0083$ & $0.1829\pm0.0093$ & $0.1765\pm0.0065$ & $0.2184\pm0.0086$ & $0.2793\pm0.0104$ & $0.2769\pm0.0054$ \\
Periodic Burgers & $\boldsymbol{0.2294\pm0.0161}$ & $0.2297\pm0.0042$ & $0.2928\pm0.0165$ & $0.3300\pm0.0564$ & $0.2463\pm0.0094$ & $0.4121\pm0.0235$ & $0.3695\pm0.0143$ & $0.3639\pm0.0063$ \\
\bottomrule
\end{tabular}

\begin{tablenotes}[flushleft]
\small

\item CNNe =  CNN-flatten embedding; Ae = Affine embedding

\end{tablenotes}
\end{threeparttable}
\end{table*}

As summarized in Table~\ref{tab:fmop_all}, FM-Operator consistently
outperforms the DeepONet baselines on all five nonlinear and
shock-dominated benchmarks. To isolate the effect of the interaction
mechanism within the FM-Operator framework from the confounding
influence of embedding (feature) differences, we compare all
interaction variants under the same CNN-flatten embedding (CNNe).
Under this controlled comparison, an MBSI-based model attains the
lowest error on every problem: FM-Operator(MBSI-CNNe) is best on
Kuramoto--Sivashinsky, square-pulse advection, and periodic Burgers,
while FM-Operator(MBSI-Ae) is best on Buckley--Leverett and cubic conservation. The
advantage of MBSI over alternative interaction mechanisms is
particularly large on tasks with sharp gradients and discontinuities: on
Kuramoto--Sivashinsky, FM-Operator(MBSI-CNNe) ($0.0155$) reduces the error by
$94.4\%$, $82.0\%$, and $46.6\%$ compared with FM-Operator(NFM-CNNe), FM-Operator(Concat-CNNe),
and FM-Operator(Outer-CNNe), respectively. These results indicate that the
bidirectional low-rank bilinear interaction of MBSI, which couples
cross-dimensional latent coordinates through complementary projection
subspaces, provides a more effective inductive bias for shock-dominated
problems than element-wise, concatenation-based, or full outer-product
interaction.

Relative to the DeepONet family, FM-Operator (MBSI) reduces the error
on Buckley--Leverett, cubic conservation, Kuramoto--Sivashinsky,
square-pulse advection, and periodic Burgers by $14.6\%$, $51.8\%$,
$88.3\%$, $39.5\%$, and $37.9\%$ over vanilla DeepONet, and by
$19.1\%$, $65.4\%$, $94.8\%$, $22.6\%$, and $44.3\%$ over
Shift-DeepONet; it also outperforms POD-DeepONet on every benchmark.

Within FM-Operator, the two embedding schemes further show that
representation construction is a task-dependent design choice. On
Buckley--Leverett and cubic conservation the affine embedding is
preferable (MBSI-Ae: $0.0845$ and $0.1019$ vs.\ MBSI-CNNe: $0.1016$
and $0.1032$), whereas on Kuramoto--Sivashinsky and square-pulse
advection the CNN-flatten embedding is better (MBSI-CNNe: $0.0155$ and
$0.1690$ vs.\ MBSI-Ae: $0.0262$ and $0.1872$). No single embedding
dominates across tasks, which is expected given the differing data
characteristics, and it underscores that representation construction
must be matched to the problem. 

\paragraph{EX2: 1D Burgers and 1D CFD Equations} 

\begin{table*}[htbp]  % 双栏文档用 table*；单栏文档请改为 table
\centering
\tiny                       % 缩小字号
\setlength{\tabcolsep}{2pt}
\begin{threeparttable}
\caption{Relative L2 error (mean $\pm$ std over 3 seeds) of FM-Operator variants and DeepONet variants.}
\label{tab:fmop_variants_exp2}
\begin{tabular}{lcccccccc}
\toprule
 & \multicolumn{5}{c}{FM-Operator} & Shift-DeepONet & Vanilla DeepONet & POD-DeepONet \\
\cmidrule(lr){2-6}
 & MBSI-CNNe & NFM-CNNe & Concat-CNNe & Outer-CNNe & NFM-Ae & & & \\
\midrule
1D Burgers & $0.1097\pm0.0022$ & $0.1040\pm0.0034$ & $\boldsymbol{0.1014\pm0.0036}$ & $0.1236\pm0.0052$ & $0.1896\pm0.0084$ & $0.2550\pm0.0151$ & $0.2769\pm0.0007$ & $0.2978\pm0.0004$ \\
1D CFD     & $0.0547\pm0.0005$ & $0.0547\pm0.0012$ & $\boldsymbol{0.0530\pm0.0011}$ & $0.0581\pm0.0011$ & $0.0705\pm0.0008$ & $0.0941\pm0.0040$ & $0.1050\pm0.0001$ & $0.1078\pm0.0004$ \\
\bottomrule
\end{tabular}

\begin{tablenotes}[flushleft]
\small
\item CNNe = CNN-flatten embedding; Ae = Affine embedding
\end{tablenotes}
\end{threeparttable}
\end{table*}

On the 1D Burgers and 1D CFD benchmarks from PDEBench (Table~\ref{tab:fmop_variants_exp2}),
all FM-Operator variants substantially outperform the DeepONet family. For the 1D Burgers case,
the best-performing variant (FM-Operator(Concat-CNNe), $0.1014$) yields error reductions of
$63.4\%$, $60.2\%$, and $65.9\%$ relative to vanilla DeepONet ($0.2769$),
Shift-DeepONet ($0.2550$), and POD-DeepONet ($0.2978$), respectively. For 1D CFD,
the corresponding error reductions are $49.5\%$, $43.7\%$, and $50.8\%$. Under the
same CNNe-based interaction ablation setup, FM-Operator(MBSI-CNNe) ($0.1097$ and $0.0547$)
remains highly competitive, trailing the best variant by merely $7.6\%$ and $3.1\%$.

The narrower performance gap between FM-Operator(MBSI-CNNe) and FM-Operator(Concat-CNNe)
observed here, in contrast to the large gap seen in EX1, can be attributed to the
considerably larger and more heterogeneous PDEBench training set ($9\,000$ instances
versus $256$ in EX1). With abundant training data comprising numerous simple and
near-steady samples, the concatenation-based variant is already capable of capturing
the necessary cross-field dependencies. The prominent performance advantage of the
FM-Operator(MBSI-CNNe) variant under the small-sample, strongly discontinuous regime
of EX1 indicates that its structured bidirectional interaction delivers a particularly
valuable inductive bias under scarce-data conditions with sharp solution features.
Finally, the embedding comparison further corroborates the task-dependent nature of
representation construction established in EX1: here the affine embedding is markedly
worse on both tasks (FM-Operator(NFM-Ae): $0.1896$ and $0.0705$ vs.\
FM-Operator(NFM-CNNe): $0.1040$ and $0.0547$), underscoring that representation
construction has a substantial, task-dependent effect on model performance.

\paragraph{EX3: Wall shear stress Prediction}

We evaluate our proposed operator network, which we refer to as
\textbf{GFM-Operator}, on the task of forecasting the transient
(pulsatile) wall shear stress (WSS) field over the entire vessel wall,
using 20 frames of sparsely sampled sensor history to predict the
subsequent 20 frames at all wall vertices. Under the corrected
8,000-sensor evaluation protocol, GFM-Operator attains a
\textbf{test relative L2 error of 0.1142} (RMSE 0.190).

Ablation of the explicit global-geometry branch reveals that injecting
the voxelized geometry representation yields \emph{no significant
improvement} over the geometry-free configuration (relative L2 error of
0.1142 with geometry versus 0.1148 without, a difference of only 0.5\%
that lies well within run-to-run noise). We argue that this is not
because geometry is uninformative for WSS prediction, but because
geometric and temporal context has already been encoded into the sensor
embeddings. Specifically, each sensor token aggregates the 20-frame
temporal trajectory of the $[\text{wall normal}, \mathrm{WSS}]$
signal through a temporal convolution, and the query-conditioned
Gaussian splatting gathers the local spatial neighbourhood around each
query point. The continuous query coordinates and surface normals
further provide local geometric cues. Consequently, the global geometry
module becomes largely redundant once such a local spatio-temporal
representation is available.

\begin{figure}[htbp]
    \centering
    \includegraphics[width=\linewidth]{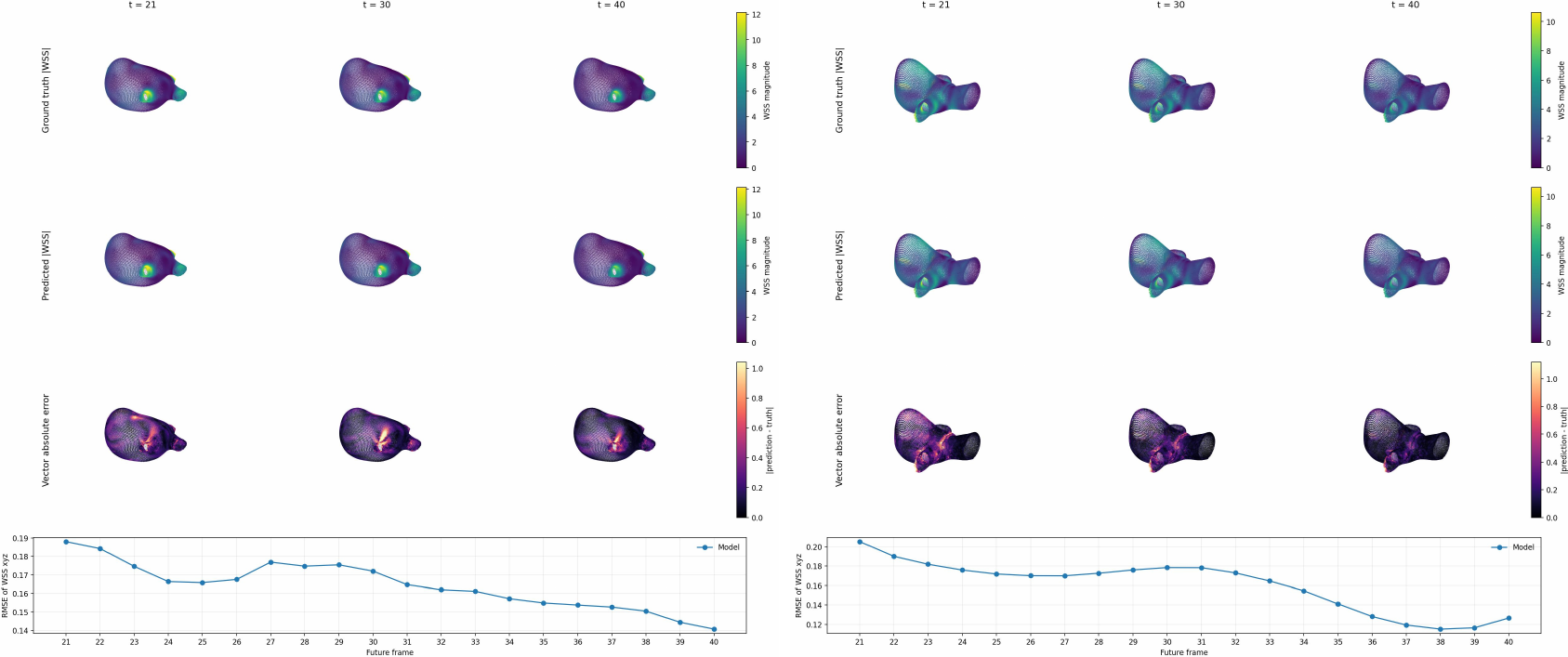}
    \caption{Representative WSS prediction results. Left: a Nuzhat case. Right: a stable case.}
    \label{fig:ex3re}
\end{figure}

Crucially, generalization to \emph{unseen morphology families}
validates that the model has internalized geometric structure rather
than memorizing the training geometries. As shown in
Figure~\ref{fig:ex3re}, the \textbf{Nuzhat deformation family}---which
is entirely absent from the training set (its 19 cases are held out as
a complete group)---exhibits prediction accuracy on par with the other
test cases, with no degradation in relative L2 error or in the visual
quality of the predicted WSS fields. Together with the case-disjoint
split over deformation families, this demonstrates that GFM-Operator
effectively learns the geometry-conditioned flow response from the
local sensor representation and transfers to vessel anatomies never
encountered during training.

\section{Conclusion and Discussion}
In this work, we revisit DeepONet and several representative variants from the perspective of feature interaction. We observe that these methods largely retain a fixed element-wise multiplicative and summation-based decoding paradigm, while their main differences lie in how the input function and query variables are embedded, conditioned, or modulated. For example, Geom-DeepONet enhances geometric awareness by introducing explicit geometric descriptors, whereas Shift-DeepONet modulates the query representation through input-dependent scaling and shifting. Motivated by this observation, we propose FM-Operator and further introduce the MBSI interaction mechanism to explicitly model richer higher-order interactions among different embedding representations. Experiments on multiple nonlinear equations with distinct solution characteristics demonstrate that the proposed approach achieves consistently competitive, and in several cases state-of-the-art, performance.

Building on this analysis, we further investigate the role of embedding design in neural operator learning. In the spatio-temporal WSS forecasting task, we construct a sensor representation that jointly encodes local geometric information and temporal history, such that each sensor embedding captures not only local field values but also spatial structure and temporal evolution. These results suggest that the performance of neural operators depends not only on the downstream interaction mechanism, but also critically on whether the input representation is appropriately aligned with the underlying physical problem.

Nevertheless, this work does not yet provide a systematic answer to a more general question: \emph{what types of embedding representations are most suitable for different classes of PDEs, and how should these embeddings interact?} Properties such as the degree of nonlinearity, solution smoothness or discontinuity, geometric dependence, and temporal scale may all influence the optimal representation and interaction mechanism. Establishing a systematic correspondence of the form
\[
\text{PDE characteristics}
\;\longrightarrow\;
\text{representation design}
\;\longrightarrow\;
\text{interaction mechanism}
\]
therefore constitutes an important direction for future research. A promising avenue is to develop a unified neural operator framework that can adaptively select, or directly learn, suitable embedding and interaction strategies according to the structural characteristics of the governing equations.

\bibliography{ref}
\bibliographystyle{tmlr}

% \appendix
% \section{Appendix}

\end{document}